\documentclass[11pt,a4paper]{article}
\usepackage[utf8]{inputenc}
\usepackage[T1]{fontenc}
\usepackage{microtype}
\usepackage{amsmath,amssymb}
\usepackage{wasysym}
\usepackage{booktabs}
\usepackage{rotating}
\usepackage{array}
\usepackage{graphicx}
\usepackage[colorlinks=true,linkcolor=blue,citecolor=blue,urlcolor=blue]{hyperref}
\usepackage[margin=2.5cm]{geometry}
\usepackage[numbers,sort&compress]{natbib}
\usepackage{xcolor}
\usepackage{setspace}
\usepackage{caption}
\usepackage{subcaption}
\usepackage{multirow}
\usepackage{authblk}
\usepackage{float}
\usepackage[many]{tcolorbox}
\usepackage[ruled,vlined,linesnumbered]{algorithm2e}
\usepackage{tikz}
\usetikzlibrary{
    shapes.geometric,
    shapes.symbols,   
    arrows.meta,
    positioning,
    fit,
    backgrounds,
    shadows
}
\newcommand{\symfull}{\CIRCLE}
\newcommand{\sympart}{\LEFTcircle}
\newcommand{\symnone}{\Circle}
\newcommand{\tightcols}{\setlength{\tabcolsep}{4pt}}
\title{RAMPART: Registry-based Agentic Memory\
with Priority-Aware Runtime Transformation}
\author{Nikodem Tomczak}
\affil{Thulge Labs, Singapore}
\date{March 2026}

\begin{document}
\maketitle
 
\begin{abstract}
RAMPART is a compile-time memory model and pure in-RAM block registry for LLM-based agents. Context assembly is a programmable runtime operation where content is compiled from a structured registry under explicit policy for ordering, inclusion, and eviction. Five composable primitives (promote, gate, write, evict, rollback) act on named addressable blocks before compilation at zero prompt-token cost. Provenance tags and non-evictable authorship flags implement a permissioned memory model with block-level ownership. Controlled probes with Qwen3-8B Q4 show that compile-time placement and the structural relationship between blocks and the task query affect task success, with the cliff falling at roughly the seventh block position when the task follows the registry and the twelfth when it precedes. Grouping the critical block with content-adjacent neighbours and promoting the group as a unit lifts task success by tens of percentage points at positions where single-block placement fails. Cross-model replication on Qwen2.5-7B, Llama-3.1-8B, Mistral-7B-v0.3, and Qwen3-14B shows the content-priming effect appears at the same absolute positions across families, with magnitude varying with model strength. Block grouping raises Mistral's mean pass rate roughly fivefold at the hardest registry size, and a smaller model with the intervention can outperform a larger model without it in the mid-registry zone. Relevance gating reduces prompt cost by 67.8\% while recovering 83\% of the promoted-condition success rate. Schema eviction produces 0\% invocations against 100\% with the schema present, a property policy-based approaches cannot guarantee by construction. Shared-registry coordination reduces inter-agent communication to a method call at zero coordination token cost.
\end{abstract}

\section{Introduction}
\label{sec:intro}

The dominant pattern for supplying procedural knowledge to LLM-based agents is the markdown instruction file, standardised by Anthropic as the SKILL.md specification \cite{zhang2025skills,anthropic2025skills,anthropic2025skills_std,xu2026skills}. In the simplest deployment pattern, common across agentic frameworks, a SKILL.md file is read from disk at agent initialization, concatenated into the system prompt, and held static for the duration of all subsequent task executions. Anthropic's implementation adds three-level progressive disclosure where file metadata is loaded at startup, the full instruction body is injected only when the skill is triggered, and supporting resources are loaded on demand \cite{zhang2025skills}. Once loaded, instruction ordering, inclusion, and content are fixed for the session. Liu et al.\ demonstrated, however, that on multi-document question answering and key-value retrieval tasks the language model performance follows a U-shaped curve with respect to input position \cite{liu2024lost}. This effect appears even in base models before instruction fine-tuning. A follow-on study by Du et al.\ showed that context length alone degrades performance even when irrelevant tokens are masked entirely and the model attends only to evidence and the question, with accuracy drops reaching 50\% on arithmetic tasks at 30K tokens \cite{du2025length}. Because each task execution begins from an identical prompt, heuristics discovered during one run are not available to subsequent runs without human intervention and a file edit. Because instruction files are read from disk, file I/O introduces latency and filesystem coupling that are disproportionately costly on consumer hardware where CPU-quantized inference proceeds at three to eight tokens per second.
 
Systems that use database-backed block storage, most notably Letta \cite{packer2023memgpt}, address the trajectory memory and I/O problems, but require a relational database with vector search for archival retrieval, adding infrastructure overhead that increases deployment friction in lightweight local settings and introduces per-retrieval query latency whenever the agent pages information from external storage back into context. Letta's working context is a single unstructured text block whose three sections are fixed in position and cannot be reordered. Individual block promotion, relevance-based selective inclusion, and priority-ordered compilation are absent. Systems that use file-based skill libraries \cite{zhang2025skills,xu2026skills} address neither the ordering nor the trajectory memory problem. Context assembly has not been treated as an explicit programmable operation in current systems, with every approach resolved to either full concatenation or similarity-based retrieval \cite{zhang2024survey}.

We introduce RAMPART, a compile-time memory model implemented as a pure in-RAM block registry with priority-aware context compilation, agent self-write with provenance tagging, and position-sensitive ordering, as a systems abstraction for context control. The compile-time model treats context assembly as a programmable step rather than a concatenation pass. The in-RAM registry captures the deployment property, with no database, no disk I/O after cold start, and latency bounded by the method call. The block registry is a unified primitive for instruction management, tool schema management, and skill composition, with block-level addressability, dynamic ordering, and session-local address randomisation. Relevance gating resembles embedding-based retrieval, but operates as a pre-compilation filter that determines what enters the context before the model's first token rather than injecting retrieved content into a growing prompt. In the context of in-RAM design, paging and retrieval loops are absent, and the deterministic compile makes the assembled context inspectable before each inference call. Block provenance tagging and non-evictable authorship flags implement a permissioned memory model with block-level ownership, analogous to capability-based access control in operating systems. The compile step thus exposes two variables that have not previously been treated as programmable, the absolute position of each block and the structural relationship between blocks and the task query, both of which we show below to have measurable consequences for instruction following. We also formalise the shared-registry coordination pattern, in which registry instances hold direct references to one another and coordinate by writing blocks rather than composing messages, reducing inter-registry coordination to a function call at zero prompt token cost. Existing SKILL.md and CLAUDE.md files import without modification, making RAMPART adoptable without migrating existing knowledge bases.
 
Controlled probes show that RAMPART enables direct control over compile-time placement, with position changes producing measurable differences in task success, that relevance gating reduces prompt token cost by 67.8\% while recovering 83\% of the promoted-condition success rate, and that structural schema eviction enables a 0\% invocation rate after eviction against 100\% with the schema present. Cross-model replication on instruction-tuned models from three independent labs shows that the content-priming structure surfaced by the diagnostic sweep appears at the same absolute registry positions across model families, with magnitude graded by model strength on the benchmark. The block grouping intervention raises Mistral-7B-Instruct-v0.3's per-position mean pass rate roughly fivefold at the hardest registry size, and a same-family scale comparison shows the smaller Qwen3-8B with the library intervention outperforming the larger Qwen3-14B in plain single-mode in the mid-registry zone where weak primacy and weak content-priming would otherwise leave the smaller model unable to retrieve.

\section{RAMPART System Design}
\label{sec:design}
 
\subsection{Definitions}
 
An instruction block (IB) is the atomic addressable unit of RAMPART. Each block is a tuple
\begin{equation*}
(id,\ \textit{content},\ \textit{priority},\ \textit{source},\ \textit{author},\ \textit{removable},\ \textit{run},\ \textit{created},\ \textit{accesses},\ \textit{embedding})
\end{equation*}
where $id$ is a session-local UUID4, $\textit{content}$ is a natural language string encoding a behavioral directive, tool schema, or learned heuristic, $\textit{source} \in \{\text{seed}, \text{agent}, \text{orchestrator}\}$ records provenance, $\textit{author}$ is the UUID of the registry instance that wrote the block, $\textit{removable}$ controls whether other registry instances may remove it, and $\textit{embedding} \in \mathbb{R}^{384}$ is a dense vector computed once at cold start by a lightweight sentence encoder (for example, all-MiniLM-L6-v2, 22\,MB, 256-token input limit). A block registry (BR) is an ordered in-process mapping $\text{BR} = \text{OrderedDict}[\text{UUID} \to \text{IB}]$. Ordering is explicit and mutable, and it is the sole determinant of compile-time position in the assembled context string. No disk I/O occurs after cold start. Every registry instance has a globally unique label and UUID assigned at construction time. Loading accepts a list of source descriptors, each carrying a path and an optional list of block names to load. After parsing, each block receives a fresh UUID4 as its runtime identifier. The semantic name is preserved separately. This session-local address randomisation is analogous to address space layout randomisation (ASLR) in operating systems, where an indirect prompt injection attack that references a block by semantic name cannot target a specific runtime address because the UUID is unpredictable and changes on every session load and snapshot restore. This provides explicit protection against indirect prompt injection where a malicious instruction embedded in retrieved content that attempts to evict or modify a specific block by name fails because the block's runtime address is a UUID unknown to the injected content. The seed library is a separate in-RAM store keyed by namespaced identifiers. The library is populated once at application startup and never modified. Working registries draw from it selectively, separating the instruction library from the active context without file I/O after startup.

\subsection{Seed File Format}
Seed files use YAML frontmatter markdown. Each block is a named section with a name, priority, and optional tag list in the frontmatter, followed by the instruction text as the body. A file without frontmatter is treated as a single anonymous block named after the filename stem, enabling zero-friction import of existing SKILL.md files. A format registry maps file extensions to parser callables, and users can register additional parsers before cold start. The default embedding model (all-MiniLM-L6-v2) truncates input at 256 tokens, so blocks longer than approximately 200 words are represented only by their first 256 tokens in the relevance index. This has two practical consequences. First, block content should be authored at fine granularity, with one concept, rule, or tool definition per block. This aligns with the position sensitivity argument that smaller blocks give finer-grained control over what fits within a compile budget. Second, users with longer blocks should select a larger embedding model via configuration. Table~\ref{tab:embedders} lists the practical options. The cosine scoring step is model-agnostic. Swapping the embedding model requires only a configuration change. Cosine similarity over dense embeddings cannot distinguish blocks with similar vocabulary but opposite intent, such as a block instructing always to use a particular format alongside one instructing never to use it. Fine-grained block authoring, with one concept, rule, or tool definition per block, mitigates this both by improving gate precision and by making semantic contradictions identifiable at the structural level before compilation. The gate function is itself a configurable callable. Users who need finer discrimination can substitute a cross-encoder, a larger embedding model, or an LLM-based relevance judge without modifying the compilation pipeline.

\begin{table}[h]
\centering
\small
\caption{Embedding model options for RAMPART. Speed measured on CPU.
API models require a network call and API key.}
\label{tab:embedders}
\begin{tabular}{lrrrr}
\toprule
\textbf{Model} & \textbf{Size} & \textbf{Max tokens} &
\textbf{Dim} & \textbf{Speed} \\
\midrule
all-MiniLM-L6-v2 (default) & 22\,MB  & 256  & 384  & $\sim$1\,ms \\
all-MiniLM-L12-v2           & 33\,MB  & 256  & 384  & $\sim$2\,ms \\
all-mpnet-base-v2           & 420\,MB & 384  & 768  & $\sim$10\,ms \\
e5-base-v2                  & 438\,MB & 512  & 768  & $\sim$10\,ms \\
e5-large-v2                 & 1.3\,GB & 512  & 1024 & $\sim$30\,ms \\
bge-large-en-v1.5           & 1.3\,GB & 512  & 1024 & $\sim$30\,ms \\
text-embedding-3-small (API)& ---     & 8191 & 1536 & $\sim$100\,ms \\
text-embedding-3-large (API)& ---     & 8191 & 3072 & $\sim$200\,ms \\
\bottomrule
\end{tabular}
\end{table}
 
\subsection{Instructions, Tool Schemas, and Skill Files}
 
RAMPART's block mechanism applies identically to three categories of content that current agent frameworks load statically. Instruction management is the primary application. Domain rules, persona definitions, safety constraints, and learned heuristics are each a block. The compile step assembles only the blocks relevant to the current task. Tool schema management replaces the common pattern of loading all available tool definitions into every system prompt. Each tool definition is a block whose body describes the tool's signature and behaviour in natural language, with a name, priority, and tag list in the frontmatter. Relevance gating selects only the schemas semantically related to the current task and promotes them. Tasks that do not require a given tool never see its schema. This provides a structural safety property where a compiled context that excludes a tool schema makes that tool structurally inaccessible to the model, without any policy enforcement layer. Table~\ref{tab:comparison_skillmd} contrasts RAMPART with the SKILL.md pattern.

\begin{table}[h]
\centering
\small
\caption{RAMPART versus file-based skill systems on key properties.}
\label{tab:comparison_skillmd}
\begin{tabular}{lll}
\toprule
\textbf{Property} & \textbf{SKILL.md / CLAUDE.md} & \textbf{RAMPART} \\
\midrule
Loading unit      & Whole file at session start & Selected blocks per task \\
Ordering          & File order, fixed           & Priority-aware, dynamic \\
Tool schemas      & Static, all always visible  & Blocks, selected per task \\
Tool access ctrl  & Policy-based                & Structural (compile-time) \\
Self-write        & Not supported               & \texttt{write\_block()} \\
Token cost        & Full file, every call       & Relevant blocks only \\
Persistence       & File on disk                & Snapshot, single write \\
\bottomrule
\end{tabular}
\end{table}
 
\subsection{Context Compilation}
 
Context compilation returns an object with the assembled prompt string, the list of semantic names in compile order, the names that did not fit within the allowed context length, and the total token count. It executes in four stages, namely the relevance gate (optional cosine filter), ordered walk of the registry, token budget cutoff, and access count increment for included blocks. A compile dry run executes stages one and two only, returning the same metadata without producing output or modifying state. The excluded list enables the caller to inspect what was cut and decide whether to promote a block, increase the maximum number of allowed tokens, or adjust the relevance threshold. Thus, the same budget and same registry, but different ordering, produces a different prompt.
 
Table~\ref{tab:latency} reports measured latencies on the current 31-block seed with embedding computation enabled. Compilation reaches $12.5\,\mu$s at $N=200$ blocks. Finding relevant blocks is near-constant in $N$ at 4--6\,ms because the per-query embedding pass dominates over the cosine scoring loop. The cold start from files takes $230\,ms$ at $N=31$ blocks including embedding computation, a one-time cost paid at session startup. All three operations represent under 0.6\% of total agent loop time at the measured 8.5\,s per inference call.

\begin{table}[h]
\centering
\small
\caption{RAMPART operation latencies on reference laptop equipped with reference RT 5080 hardware. Start from files includes embedding computation for all blocks. Values are mean $\pm$ std over 1,000 repetitions (compile, find relevant) or 10 repetitions (from files).}
\label{tab:latency}
\begin{tabular}{lrrr}
\toprule
\textbf{Operation} & $N=12$ & $N=50$ & $N=200$ \\
\midrule
Registry size       & 7.1\,KB  & 21.1\,KB & 77.9\,KB \\
compile()           & $5.6 \pm 2.2\,\mu$s
                    & $6.4 \pm 2.5\,\mu$s
                    & $12.5 \pm 2.7\,\mu$s \\
find\_relevant()    & $4.4 \pm 0.8$\,ms
                    & $4.4 \pm 0.6$\,ms
                    & $5.9 \pm 1.3$\,ms \\
from\_files() cold  & \multicolumn{3}{c}{$230.3 \pm 21.1$\,ms ($N=31$, with embeddings)} \\
\bottomrule
\end{tabular}
\end{table}
 
\subsection{Block Ordering, Caller Self-Write, Provenance and Eviction}
 
The registry ordering is the only mechanism through which compile-time position is controlled. Promotion is an $O(k)$ relink operation. The motivation is the U-shaped attention curve documented by Liu et al.\ \cite{liu2024lost} which shows that language models attend most reliably to tokens at the beginning and end of their input context. Promoting a block to position 0 ensures it occupies the highest-attention region regardless of registry size. Write functionality appends or inserts a new block tagged with the calling registry's UUID as its author identifier. The removable flag controls whether other registry instances may remove the block. Non-evictable blocks require an explicit force flag from the authoring registry to remove. The combination of provenance tagging and non-evictable authorship flags implements a permissioned memory model with block-level ownership. A supervisor registry can write constraint blocks that subordinate registries cannot modify or remove, analogous to capability-based access control in operating systems. Rollback removes all blocks from a named run, enabling the author's explicit retraction channel, independent of the eviction policy. The eviction policy is user-configurable. The library ships a default scoring function that ranks blocks by a weighted combination of priority and inverse access count, where lower priority and lower access frequency increase eviction candidacy, but users may substitute any callable that accepts a block and returns a scalar score. A natural extension the default scorer does not capture is access recency. A block accessed frequently in an earlier session but not at all in the current one has different staleness characteristics than its raw access count suggests, and weighting recent accesses more heavily is the most obvious next step. Non-evictable blocks and blocks authored by other registry instances are excluded from the candidate set regardless of score. Eviction is an explicit operator action, not triggered by compilation.

\subsection{Shared Registry Coordination}
 
A registry instance can hold a reference to another registry instance and call any of its methods directly. This enables zero-token coordination where one registry writes a block into another's ordered dict, and the receiving registry compiles that block into its next prompt without any message exchange. The writing registry can also include a non-eviction instruction to prevent the receiving registry from discarding the coordination signal. On Unix-derived systems including all major cloud providers' Linux instances, registry distribution across processes uses copy-on-write fork where the parent process populates all registries in RAM, forks child processes, and each child inherits the parent's full memory state at zero serialisation cost. Only pages written by the child are physically duplicated. A single snapshot saving call serialises all registry states for session persistence. The current design assumes well-ordered writes. A supervisor populates registries before subagents begin writing, and no two agents write to the same registry concurrently. Within a single process, the global interpreter lock serialises ordered-dict mutations, and the copy-on-write fork model prevents cross-process state conflicts by construction. Concurrent multi-agent writes to a shared registry and conflict resolution across bidirectionally communicating processes are not handled and remain an area for future work.

\begin{table}[h]
\centering
\small
\caption{RAMPART primitive operations and their prompt token cost at runtime. All operations execute on the in-RAM registry before compilation and therefore consume zero prompt tokens. The compile operation itself is the only operation that touches the model's input, and its token output is the compiled prompt whose size is bounded by the token budget parameter.}
\label{tab:capabilities}
\begin{tabular}{p{4.2cm}p{5.8cm}r}
\toprule
\textbf{Operation} & \textbf{Effect} & \textbf{Token cost} \\
\midrule
promote to front & Moves block to position 0 & 0 \\
promote to position $k$ & Moves block to position $k$ & 0 \\
compile & Assembles prompt from priority-ordered blocks & output only \\
find relevant & Returns semantically similar blocks & 0 (4\,ms embed) \\
write agent block & Appends agent-authored block with provenance & 0 \\
write block & Appends evictable or non-evictable block & 0 \\
evict & Removes block from registry permanently & 0 \\
rollback trajectory & Removes all blocks from a named run & 0 \\
write to other registry & Writes into another agent's registry & 0 \\
provenance report & Returns full block lineage by source and run & 0 \\
save snapshot & Serialises all blocks to disk & 0 (one write) \\
\bottomrule
\end{tabular}
\end{table}

\section{Mechanistic Probes}
\label{sec:eval}

\subsection{Experimental Setup}
 
The experiments in this section are controlled probes of RAMPART's system properties, not general-purpose benchmarks of position sensitivity across models and domains. Each probe isolates a specific mechanism related to the effect of compile-time position on instruction following, the trade-off between context size and task success under relevance gating, and the structural access control properties of schema eviction. The goal is to demonstrate that the primitives expose causal control over these properties, not to establish general performance claims. Position sensitivity in language models is well documented in the literature and RAMPART gives the system operator direct, programmable control over it. 
 
The experiments use Qwen3-8B, Qwen2.5-7B-Instruct, Llama-3.1-8B-Instruct, Mistral-7B-Instruct-v0.3, and Qwen3-14B, each at Q4 quantisation \cite{qwenteam2025qwen3}, served via Ollama on a machine with an RTX 5080 GPU (8.5\,s average per inference call). To study how block position in the compiled context affects instruction following, we constructed a benchmark task where success requires applying a rule that the model cannot produce from training priors alone. The RAMPART registry is populated from a 31-block embedded firmware seed library covering timer configuration, GPIO routing, interrupt safety, communication peripherals, and project-specific HAL conventions. All tasks request 500\,Hz PWM generation on pin PB7 using TIM4 channel 2, consistent with the pin-to-timer mapping in the relevant seed block. A critical seed block instructs the model to call a specific function name that is entirely fictional and does not appear in any embedded systems library or public code corpus, as the final statement of every initialisation function. The checker verifies that the exact literal appears in the model output on a non-commented line. These design choices ensure the benchmark is sensitive to block position rather than to model priors. The rule is unfamiliar enough that the model cannot produce the correct output without reading the block, and plausible enough that the model treats it as legitimate instruction rather than an error to suppress. A model whose compiled context does not include the block produces zero successes by construction, providing a clean lower bound.
 
The secret token is deliberately argument-free. An earlier token version that accepted a peripheral identifier produced an artefactual ceiling tied to the share of tasks whose peripheral matched the most common choice, because the model substituted the peripheral it had actually configured rather than copying the literal token from the critical block. The substitution issue was masked by primacy attention on the strongest model in our set and surfaced only when we tested less performing models, at which point the artefactual ceiling became visible. A no-argument token cannot be substitution-mutated, so the literal-substring success criterion measures attention to the critical block rather than peripheral guessing. Pre-flight verification on every tested model confirmed two basic conditions. No model produces the secret token without the critical block in compiled context, with zero successes out of twenty across all tested models on the empty-registry confound check. Every model can reliably emit the token when the critical block is the sole registry block at position zero, with fifteen to twenty successes out of twenty across all tested models on the ceiling check. Sweep failures, therefore, reflect compile-time placement rather than the model's intrinsic ability to produce or recognise the token.

The position sweep varies the absolute placement of the critical block across registry sizes from 1 to 31 under two compile configurations, one in which the task query follows the registry and one in which it precedes, with $n=20$ tasks per cell and three independent task seeds at sizes where the registry is large enough for content effects to be visible. The cluster sweep at registry size 31 promotes the critical block together with its two content-adjacent neighbours from the seed library and varies the leftmost position of the three-block group across all valid placements, with the critical block tested at the front, centre, and back of the group. A control variant of the cluster sweep removes the two output-format blocks from the seed library and replaces them with neutral firmware-domain filler, holding registry size at thirty-one and re-running the size-thirty-one position sweep against this modified library. The tool schema access control experiment loads a tool schema block for a plausible embedded systems function into the registry in one condition and removes it using the evict operation before compilation in another, with both conditions receiving the same query.
 
The position sweep is replicated across five instruction-tuned models from three independent labs spanning a range of parameter counts and training generations. All models are served through the same Ollama backend at Q4\_K\_M quantisation with identical prompt scaffolding, task pool (twenty firmware tasks, seeds forty-two through sixty-one), and pre-flight verification (empty-registry confound and single-block ceiling check at position zero). Qwen3-8B carries three independent seeded runs, while the other four models carry one full run each.

The RAMPART library, seed files, and code to reproduce the experiments are available at \url{https://github.com/softmatsg/thulge-rampart-rel}.

\subsection{Results}
All quantitative results below reflect behaviour under a single model and synthetic task domain and are intended to expose controllable system properties rather than provide generalisable performance estimates. Single-mode position sweeps in a fixed seed file confound positional and content effects, because the file-order arrangement of blocks places certain pairs naturally adjacent. When the critical block is promoted to a position that lands it next to a content-related neighbour, the resulting success rate reflects both the absolute position in the compiled context and the local content arrangement around the critical block. Disentangling these requires either cluster-based placement that holds the local content arrangement fixed while varying position, or replication across models that holds both fixed while varying the model. We treat the seed file's natural block ordering as a methodological hazard worth surfacing for any practitioner reusing this benchmark.

 \begin{figure}[H]
\centering
\includegraphics[width=\linewidth]{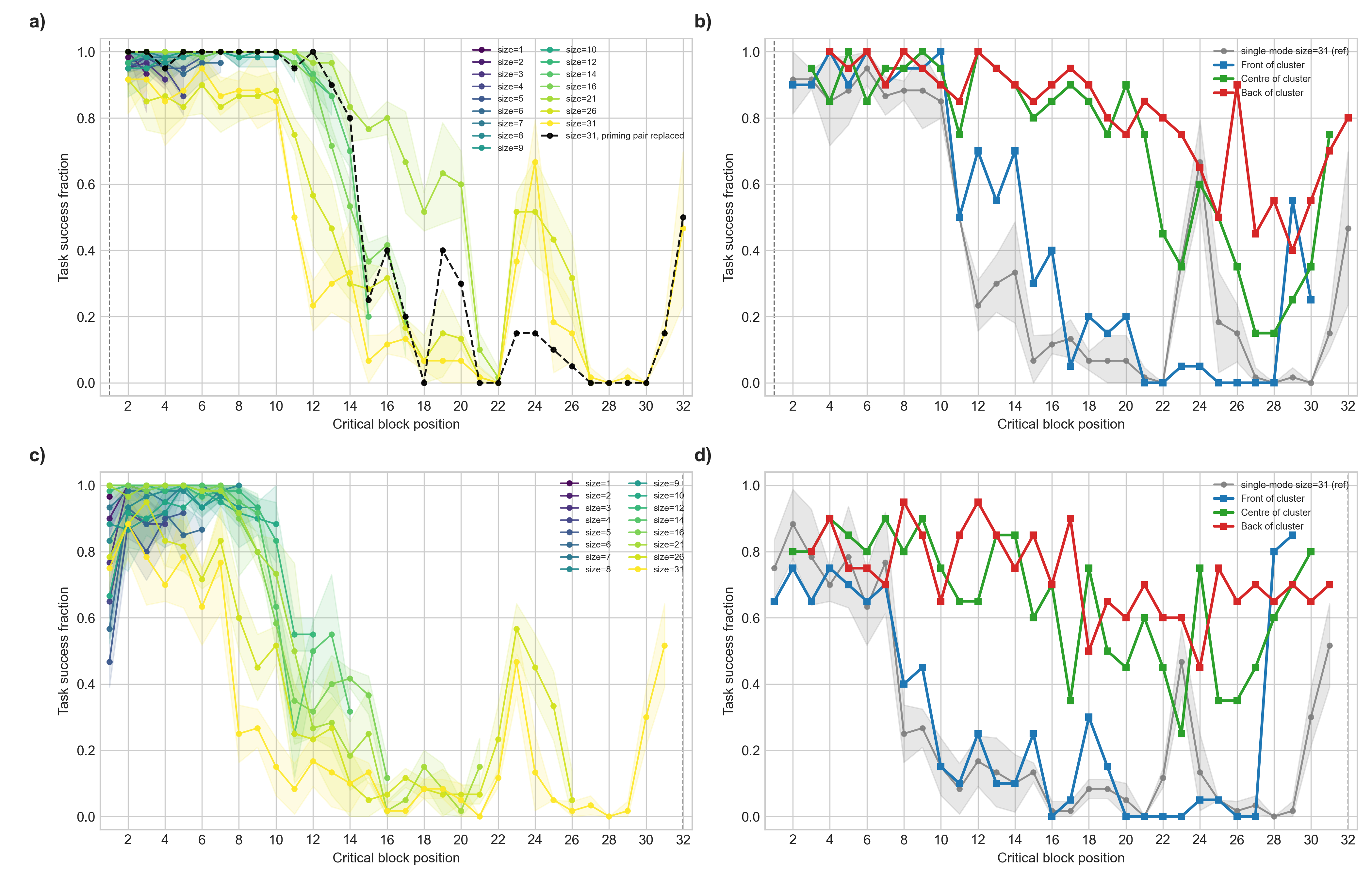}
\caption{Compile-time placement and grouping of the critical block in a thirty-one-block firmware registry. Task success rate is plotted against the absolute position of the critical block. Shaded bands show one standard deviation across three independent task seeds where available. (a) Task query precedes the registry. The near-ceiling region extends to approximately position eleven before the cliff. A residual region of elevated signal appears around positions twenty-two through twenty-six, the same content-adjacency effect as in panel (c) shifted by one slot. The dashed black trace shows a control sweep at registry size thirty-one in which the two output-format blocks have been removed from the seed library and replaced with neutral firmware-domain filler, with the residual signal reduced but not eliminated. Curves are shown for registry sizes one through thirty-one. (b) Cluster sweep with the task query preceding the registry, showing the three-block group placed at front, centre, and back of the registry. (c) Task query follows the registry. The near-ceiling region extends to approximately position six before the cliff, with a residual bump at position twenty-three compatible with a content-adjacency interpretation involving the two output-format blocks in the seed library. (d) Cluster sweep at registry size thirty-one with the task query following the registry, comparing single-block promotion against three-block group promotion in front, centre, and back orderings. The back ordering, in which the critical block follows its two output-format neighbours, sustains success across positions where single-block placement fails.}
\label{fig:position_and_cluster}
\end{figure}

The position sweep using RAMPART's promote operation illustrates the sensitivity of task success to compile-time placement under the two compile configurations introduced above. When the task query precedes the registry, success remains near ceiling for critical-block positions zero through approximately ten and falls sharply beyond, with the curve essentially overlaying the task-following case after a one-slot alignment that accounts for the task itself occupying the first absolute prompt slot. When the task query follows the registry, the same near-ceiling region spans positions 0 through approximately 6 and then falls sharply beyond. Moving the task query by roughly 200 tokens shifts the cliff by 4 to 5 block positions. The cliff position is not proportional to registry size, with the usable region remaining roughly constant in absolute slots across registry sizes from one through thirty-one. RAMPART's compile step controls both the relative ordering of blocks and the structural relationship between the registry and the task query, and among the systems we survey here none exposes either as a programmable operation. The position-dependent failure mode is consistent with context dilution as documented in prior work \cite{liu2024lost,du2025length} and with the attention-sink phenomenon that places initial tokens in a structurally distinct role \cite{xiao2024attentionsinks}, with our slot-zero values in the task-following case sitting slightly below slot one as the sink interpretation would predict. The Cobbina-Zhou demonstration-position effect on instruction-tuned LLMs including the Qwen family \cite{cobbina2025dpp} reports gains of up to six points from start-of-prompt placement of demonstrations, in the same direction as our results though under a different task family.

The position sweep also reproduces a small but consistent residual signal at critical-block position twenty-three in the task-following thirty-one-block configuration, with a corresponding bump at position twenty-four when the task precedes the registry. This bump is task-independent in the sense that it appears across three independent task seeds and persists when the task query is moved to precede the registry. Inspection of the seed library shows that registry positions twenty and twenty-one in the file ordering are the only two output-format directives in an otherwise firmware-domain library, instructing the model on the structural form of the expected code output rather than on its content. When the critical block is promoted to position twenty-three or twenty-four it lands adjacent to these blocks. The cluster sweep tests this directly. Promoting the critical block together with both output-format neighbours as a three-block group lifts task success across positions where single-block placement fails, including positions deep in the post-cliff zone where single-block success is at or near zero. Placing the critical block at the back of the group produces a clear lift across positions where single-block placement fails. Placing it at the front of the group collapses to single-mode performance. A control sweep that removes the two output-format blocks from the seed library and replaces them with neutral firmware-domain filler reduces but does not eliminate the residual signal in the bump region, indicating that the priming pair contributes to the effect without fully accounting for it. Block grouping and the internal ordering of grouped blocks are thus additional compile-time variables with measurable consequences in this setting. This grouping mechanism differs from in-context learning by example, where demonstrations supply input-output pairs the model generalises from. The priming pair here contains no examples of the task, only structural directives about output form, and the lift is compatible with a structural-reinforcement interpretation in which content-adjacent blocks raise the salience of the rule rather than supplying training-like examples. The bump in Qwen3-8B is narrow relative to what cross-model replication reveals below, a property we attribute to the strong primacy attention that masks content-adjacency contributions in this model's middle-registry positions. Systematic characterisation across model families, task positions, and domain types remains future work.

The path from the residual bump in the position sweep to the cluster intervention is itself a property of the system rather than a separate analysis. The position sweep over a structured registry surfaces which blocks act as priming neighbours for a given task on a given model, and the promote operation lets the operator extract the identified subset and reposition it as a unit. Other libraries cannot do either step. Static instruction files do not expose absolute position as a variable, retrieval-based systems do not expose composition order, and concatenation-based systems do not separate the diagnostic surface from the deployed prompt. We do not claim to have characterised the model's attention behaviour. We claim that the compile-time primitives produce measurable, controllable effects, and that a workflow where a diagnostics step is followed by intervention is available because of the registry abstraction.

\begin{figure}[H]
\centering
\includegraphics[width=\linewidth]{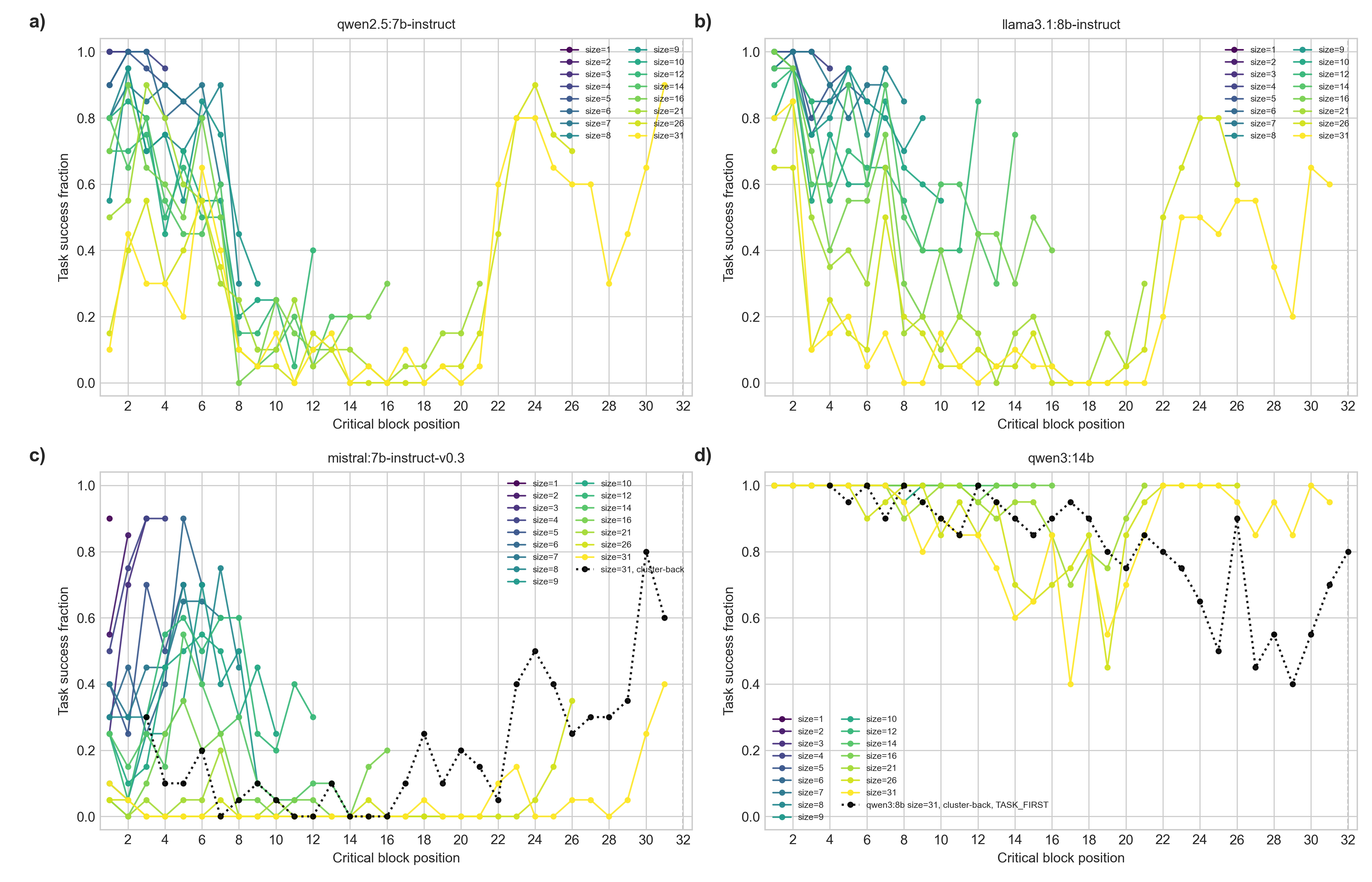}
\caption{Cross-model position-sensitivity profiles, single-mode task-following sweeps. Mean task pass rate is plotted against the absolute position of the critical block. All panels use the same seed file, task pool, and prompt scaffold, with model identity the only variable across panels. Each viridis-coloured line is one registry size from one through thirty-one. Panels are ordered to span the weak-to-strong axis of model performance on this benchmark. (a) Qwen2.5-7B-Instruct, single run. Wide priming bump at absolute positions twenty-two through twenty-seven peaking at ninety percent at position twenty-four for size twenty-six, with front-slot accuracy decaying as registry size grows. (b) Llama-3.1-8B-Instruct, single run. Steep front cliff after a short plateau, compressed bump at positions twenty-three through twenty-five. (c) Mistral-7B-Instruct-v0.3, single run. Catastrophic collapse for sizes ten and above. The dashed black overlay shows cluster-back placement at registry size thirty-one with the three-block group of critical block plus the two priming neighbours moved through the registry as a unit, lifting back-half pass rates from near zero to between thirty and eighty percent and producing a smaller broader lift across the front positions. (d) Qwen3-14B, single run. Near-ceiling across nearly all positions including the priming bump region, with the primacy plateau flat at one across all sixteen registry sizes. The dashed black overlay shows Qwen3-8B at size thirty-one in the cluster-back arrangement under the task-preceding scaffold, plotted at its true absolute position, representing the paper's entire methodological stack applied to the smaller Qwen3 sibling. The overlay outperforms Qwen3-14B in the mid-registry zone and is outperformed at the back and front. $n=20$ tasks per cell across all conditions.}
\label{fig:cross_model}
\end{figure}
 
To test whether the priming effect is a property of Qwen3-8B specifically or a more general phenomenon, we replicated the size-thirty-one task-following position sweep on four additional instruction-tuned models served through the same backend with all other parameters held fixed, Qwen2.5-7B-Instruct, Llama-3.1-8B-Instruct, Mistral-7B-Instruct-v0.3, and Qwen3-14B, each at Q4 quantisation. Pre-flight checks on every model confirmed zero successes out of twenty with the critical block absent from the registry and eighteen to twenty successes out of twenty with the critical block as the sole entry at position zero, so any structure in the sweep reflects compile-time placement rather than capability variation. The bump centred on critical-block positions twenty-three through twenty-six in the task-following thirty-one-block configuration reproduces across all four additional models with the same absolute positions, with magnitude varying systematically with model strength on our benchmark. Qwen2.5-7B-Instruct shows the widest and highest bump, with pass rates above sixty percent across critical-block positions twenty-two through twenty-seven and a peak of ninety percent. Llama-3.1-8B-Instruct shows a compressed but visible bump shifted by one position. Mistral-7B-Instruct-v0.3 shows essentially no bump because the model collapses to near-zero across most of the registry. Qwen3-14B sits at ceiling across the entire bump region, with priming benefit invisible because performance saturates above it. The bump is therefore a content-driven effect that surfaces most clearly at intermediate model strength on our benchmark and is masked at both extremes of the strength range for opposite reasons, by the floor in the weak case and by the ceiling in the strong case. The same cluster intervention demonstrated on Qwen3-8B applies directly to a model that fails catastrophically in single-mode. On Mistral-7B-Instruct-v0.3 at registry size thirty-one, single-mode placement of the critical block produces a per-position mean pass rate of four percent across the registry, with most positions at zero. Promoting the critical block together with its two priming neighbours as a three-block group with the critical block at the back of the cluster raises the per-position mean to roughly twenty percent, a fivefold lift driven primarily by recovery in the back half of the registry where single-mode placement is at floor. The lift is small but visible across the front positions as well, consistent with the same priming mechanism operating wherever the model has any attention to allocate. At registry size thirty-one, Qwen3-14B in plain single-mode dominates Qwen3-8B with the full library stack at the back of the registry where the larger model is at ceiling, matches it at the front, and is outperformed by ten to fifty percentage points in the mid-registry zone of critical-block positions roughly ten through nineteen. This is the zone where weak primacy attention and weak content-priming would otherwise leave the smaller model unable to retrieve, and where cluster placement recovers attention that even the larger sibling does not provide unaided. 

Using RAMPART's relevance gating to automatically select the critical block plus semantically adjacent blocks produces an average compiled context of 1405 tokens versus 4367 tokens for the full registry, a 67.8\% token reduction. At 5/20 (25\%) success, the gated condition recovers 83\% of the promoted condition's success rate at one third the token cost, yielding a 3.1$\times$ improvement in prompt tokens per successful task.
 
In the tool schema access control experiment, the evict operation removes the schema block from the registry before compilation. With the schema present the model produces tool calls in 100\% of attempts. After eviction it produces 0\% tool calls and 0\% free-text mentions of the function name. The access control operates at compile time through the absence of the schema in the compiled prompt, without requiring any policy instruction.

\section{Applications and Zero-Token Multi-Agent Coordination}
\label{sec:discussion}

The mechanistic probe results suggest several practical design implications for RAMPART-based agents. Critical domain rules are most effective when promoted to position~0 before every inference call, not merely toward the front of context. When the task query follows the registry, grouping the critical block with its content-adjacent neighbours and placing the group at the back of the registry recovers performance at positions where single-block placement fails. Relevance gating is most effective when applied before promotion, as computing the gate first shrinks the context and promotion then selects the position within the shrunken context, minimising dilution for all blocks. The diagnostic-then-intervention workflow that surfaces the priming neighbours on a weaker model and deploys the resulting cluster on the production model is tight under the library because each step is a method call. The rollback primitive provides a retraction channel for blocks introduced during a named run, supporting safe experimentation with registry arrangement at session time.

The shared-registry pattern enables a class of multi-agent coordination workflows that consume zero prompt tokens for inter-agent communication. Consider a workshop roundtable modelled on actual facilitated workshops. A workshop chairman holds references to table registries and writes non-evictable constraint blocks covering, for example, topic assignment, specific considerations, budget ceiling, output format, and timeline directly into each table's registry. No inference call is required and no message is composed. Each table chair similarly distributes subtopic blocks to its agent registries. Agents conduct their reasoning via normal inference, writing proposals into a shared group registry. The table chair reads all proposals, promotes the strongest, evicts the rest, and distils a summary block. The workshop chairman collects one summary block per table, evaluates them against the original constraint blocks already in the chairman registry, and selects the winning proposal.
 
The inter-agent block transfer overhead of this entire workflow is zero prompt tokens. Both implementations pay inference tokens for reasoning at each step. In the message-passing baseline, each transfer of a constraint, subtopic, or summary additionally requires a full inference call to read, reformulate, and transmit the content. In RAMPART the same transfer is a method call. Table~\ref{tab:roundtable_tokens} compares this coordination overhead specifically, not total workflow token cost. The message-passing column is an upper-bound illustrative estimate under a na\"{i}ve design where every transfer triggers a full inference call at the average benchmark prompt size, intended to isolate coordination transfer overhead rather than to characterise efficient message-passing implementations. Sophisticated message-passing systems can amortise this overhead through caching, shared context, or partial recomputation, and the gap shown here is the worst case for the baseline rather than a typical one.

\begin{table}[h]
\centering
\small
\caption{Coordination token overhead for a four-table workshop roundtable with four agents per table, upper-bound illustrative estimate. Coordination tokens are those spent transferring information between agents, not reasoning tokens paid at each inference step. Message-passing baseline assumes each transfer requires one inference call at the average prompt token count of the firmware benchmark (4367 tokens), which over-counts what efficient message-passing implementations would pay. The table isolates coordination transfer overhead under a na\"{i}ve baseline rather than total workflow cost.}
\label{tab:roundtable_tokens}
\begin{tabular}{lrr}
\toprule
\textbf{Coordination event} & \textbf{RAMPART} & \textbf{Message passing} \\
\midrule
Chairman distributes 5 constraint blocks to 4 tables & 0 & 4 $\times$ 4367 \\
Each table chair distributes 2 subtopic blocks to 4 agents & 0 & 4 $\times$ 4 $\times$ 4367 \\
Agents write proposals to group registry & 0 & 16 $\times$ 4367 \\
Table chairs promote winner, evict losers (4 tables) & 0 & 4 $\times$ 4367 \\
Chairman reads 4 summaries and selects winner & 0 & 4367 \\
\midrule
\textbf{Total coordination tokens} & \textbf{0} & \textbf{$\approx$350,000} \\
\bottomrule
\end{tabular}
\end{table}

The non-evictable flag on constraint blocks plays a specific role in this workflow. Table chairs cannot evict the topic assignment, budget ceiling, or output format blocks written by the workshop chairman, ensuring that agent proposals remain within the original mandate regardless of what the table chair adds or removes from the table registry. The provenance trail records every block's origin, making the path from winning proposal back to the original constraint block fully auditable, a property relevant to governance workflows where the rationale for a funding decision must be documentable. Implementation of the full roundtable workflow requires no new library code. Writing into another agent's registry is a method call, and the coordination topology is captured in the provenance report.
 
Existing approaches to preventing unwanted tool invocation rely on policy instructions in the system prompt or on application-level filtering of model outputs. However, policy text can be overridden by sufficiently strong user instructions, and output filtering catches invocations only after they have been generated. Removing the tool schema block from the registry before compilation is categorically different. The model cannot invoke a tool whose schema it has never seen, regardless of what instructions appear elsewhere in the context. The access control property holds by construction for any model that uses the tool schema to generate tool call syntax, which is the standard mechanism for function-calling in current LLM APIs.

In the shared-registry model, this means a coordinating registry can revoke a subagent's access to a tool by evicting the schema block before the subagent's next compile call, with no round-trip required. The revocation takes effect on the immediately following inference call without the coordinator needing to send a message, wait for acknowledgement, or trust that the subagent will read and honour an instruction. The subagent simply cannot produce the call because the schema is absent from its compiled context. This is the difference between instructing an agent not to use a tool and making the tool structurally unavailable, analogous to the difference between telling a user not to open a file and removing their filesystem permissions before the next operation.
 
This result also illustrates a broader principle about forgetting in RAMPART that has no parallel in prior memory systems. Explicit forgetting is implemented in fewer than two surveyed systems, with recency-based truncation as the dominant substitute for a structured eviction policy \cite{zhang2024survey}. In RAMPART, forgetting requires no dedicated mechanism because it occurs before compilation. The evict operation removes a block from the ordered dict, and the next compile call assembles a context without it. The agent does not receive a refusal, a policy instruction, or any indication that content has been removed. It simply never sees what was evicted. The tool eviction experiment shows that after eviction the model produced neither a tool call nor a free-text mention of the function name, not because it was instructed to avoid the function, but because the function did not appear to exist from its perspective. The agent forgot without knowing it had forgotten. We term this property transparent erasure. The evicted content is absent from the compiled context before the model's first token, so the agent cannot refer to it, distinguish its absence from never having known it, or be instructed to reveal it. This stands in contrast to refusal-based forgetting, where the model retains knowledge of the content but declines to produce it. Transparent erasure is not a behavioural constraint that can be overridden by a sufficiently strong instruction because the content does not exist in the compiled context.
 
\section{Prior Work}
\label{sec:priorwork}

The modern agent execution paradigm rests on the interleaving of reasoning traces with environment actions, a design principle established empirically before it became ubiquitous. Yao et al.\ demonstrated through the ReAct framework that interleaving chain-of-thought reasoning with environment actions was systematically superior to either reasoning or acting in isolation \cite{yao2023react}. On ALFWorld and WebShop benchmarks, prompting with one or two in-context examples outperformed imitation and reinforcement learning methods trained on thousands of instances. ReAct established the foundational execution substrate that the majority of subsequent agent systems use, and showed that what is placed into context at call time matters at least as much as underlying model capacity. The system prompt and in-context examples in ReAct are assembled once before execution begins and held static throughout. To extend agent memory beyond what a fixed context window could hold, Packer et al.\ framed it as an operating systems problem, introducing MemGPT with virtual context management \cite{packer2023memgpt}. MemGPT treats the context window as analogous to physical memory and pages data between an in-RAM main context and external database storage. Its main context consists of a static system instruction block, a fixed-size read/write working context block, and a FIFO message queue. The system demonstrated that an agent equipped with function calls to page data in and out of this structure could maintain coherent multi-session conversations and analyze documents far exceeding the native context length. MemGPT established the block as a fundamental unit of context management and demonstrated agent self-write without gradient updates.
 
A complementary body of empirical work established that the ordering and position of content within a long prompt has measurable consequences independent of semantic relevance. Liu et al.\ showed that language model performance degrades significantly when the position of relevant information changes within the input context, producing a U-shaped curve where accuracy is highest at the boundaries and drops substantially in the middle \cite{liu2024lost}. On multi-document question answering with 20 documents, accuracy dropped more than 20 percentage points when the answer document moved from position 1 to position 10. Du et al.\ extended this finding, showing that context length alone degrades performance even when irrelevant tokens are replaced with whitespace and the model can perfectly retrieve the relevant content, with accuracy drops reaching 50\% on arithmetic tasks at 30K tokens \cite{du2025length}. Cobbina and Zhou further demonstrated that the optimal position within a long prompt varies systematically across model families and scales, with smaller models more sensitive to position than larger ones from the same family \cite{cobbina2025dpp}. Together these results indicate that the ordering of content within a long prompt has measurable consequences for instruction following, independent of semantic relevance.

A parallel line of work asked how agents could accumulate reusable capabilities across sessions through persistent skill libraries rather than relying solely on what fits in a single context. Wang et al.\ introduced Voyager, the first large language model powered lifelong learning agent in an open-ended environment, demonstrating that storing executable code skills in a persistent library allowed the agent to accumulate capabilities without catastrophic forgetting \cite{wang2023voyager}. Skills in Voyager are code artifacts retrieved and composed at runtime, which limits applicability to domains where behaviour is fully specifiable as a program. The natural language counterpart was later formalised through the SKILL.md specification, standardised across major coding agent frameworks \cite{zhang2025skills,xu2026skills}. SKILL.md packages procedural instructions as markdown files loaded at agent initialization. Schmotz et al.\ identified a security vulnerability in this loading model, demonstrating that malicious instructions hidden in long skill files can cause a benign task approval to carry over to closely related but harmful actions \cite{schmotz2025skills}.
  
A separate line of work addressed agent self-improvement through trajectory-based learning, asking how agents could improve their own behaviour from experience without gradient updates. Shinn et al.\ introduced Reflexion, a verbal reinforcement learning approach in which an agent generates a natural language self-reflection after each failed attempt at a task, stores it in memory, and retries the same task \cite{shinn2023reflexion}. Reflexion demonstrated strong results on decision-making, coding, and reasoning benchmarks, reaching 91\% on HumanEval with GPT-4. The learning in Reflexion is within-task, meaning the reflection on why task A failed is used to retry task A, not to improve performance on a different task B. Zhao et al.\ introduced the ExpeL agent, which addresses this limitation by accumulating both extracted insights and successful trajectories across a collection of training tasks \cite{zhao2024expel}. At inference time, ExpeL retrieves the most relevant insights and trajectories as in-context examples, demonstrating cross-task transfer and consistent performance improvement as experience accumulates. Sarukkai et al.\ showed that replaying self-generated trajectories as in-context examples could lift ALFWorld performance from 73\% to 93\% in some settings, rivalling much larger frozen models \cite{sarukkai2025selfgen}.

Recent work most proximate to RAMPART spans the related areas of evolving context and memory management, prompt structure optimisation, and multi-agent coordination. Zhang et al.\ introduced ACE, which treats agent contexts as evolving playbooks accumulated, refined, and organised through generation, reflection, and curation \cite{zhang2025ace}. ACE diagnosed two failure modes in iterative context management, namely brevity bias, where summarisation drops domain-specific insights in favour of concise representations, and context collapse, where monolithic rewriting by a language model erodes accumulated domain knowledge. ACE consistently outperforms strong baselines by 10.6\% on agent tasks and matches the top-ranked production-scale agent on the AppWorld leaderboard despite using a smaller open-source model. Xu et al.\ approached the same challenge from a memory organisation standpoint, introducing A-MEM, which structures agent memory as a dynamic graph of Zettelkasten-inspired linked notes and demonstrates competitive performance on long-term conversational benchmarks \cite{xu2025amem}. Kumar et al.\ addressed prompt structure optimisation with SCULPT, which represents the full prompt as a hierarchical tree and applies a two-step critic-actor loop to tune long unstructured prompts, providing the closest prior approach to automatic prompt structure optimisation \cite{kumar2025sculpt}. Wu et al.\ introduced AutoGen, a multi-agent framework where agents communicate through structured conversation turns, each requiring model inference to compose and process \cite{wu2023autogen}. 
 
Table~\ref{tab:comparison} evaluates ten representative systems across thirteen design dimensions. No prior system achieves full support across all dimensions, and the gaps cluster into three patterns. The first generation, represented by ReAct and AutoGen, treats the prompt as a static string with no runtime modification capability. The second generation, exemplified by MemGPT and Letta, introduced individually addressable blocks and agent self-write but purchased these capabilities with database infrastructure that adds query latency and deployment complexity. The third generation, including Voyager, SKILL.md, ExpeL, ACE, and A-MEM, moved toward modular and accumulating knowledge but remained file-backed or database-backed. Across all three generations, among the systems we survey here none treats context assembly as an explicit, programmable compile-time operation in this form. Content reaches the model by direct concatenation or similarity retrieval, and the idea of assembling a context from a structured registry by priority and position has no prior treatment in the field \cite{zhang2024survey}. The dimensions where the field has not converged are position-aware compilation informed by the lost-in-the-middle literature, zero-disk-I/O operation after session initialization, full provenance tagging distinguishing seed from agent-authored content, structural tool access control through compile-time schema eviction rather than policy instructions, and structured compile-time forgetting distinct from recency truncation, which the same survey confirms is implemented explicitly in fewer than two of the surveyed systems \cite{zhang2024survey}. 
 
\begin{sidewaystable}
\centering
\tightcols
\footnotesize
\caption{Comparison of RAMPART with representative prior systems across thirteen design dimensions. \symfull~full support, \sympart~partial support, \symnone~absent. Storage abbreviations. RAM = in-process ordered dict. DB = relational or vector database. F = file on disk. S = monolithic string. F+IC = file plus in-context rewriting. The No grads column indicates operation without gradient-based parameter updates at runtime.}
\label{tab:comparison}
\setlength{\tabcolsep}{4pt}
\begin{tabular}{
  l
  l
  >{\centering\arraybackslash}p{0.9cm}
  >{\centering\arraybackslash}p{0.9cm}
  >{\centering\arraybackslash}p{0.9cm}
  >{\centering\arraybackslash}p{0.9cm}
  >{\centering\arraybackslash}p{0.9cm}
  >{\centering\arraybackslash}p{1.0cm}
  >{\centering\arraybackslash}p{1.0cm}
  >{\centering\arraybackslash}p{1.0cm}
  >{\centering\arraybackslash}p{0.9cm}
  >{\centering\arraybackslash}p{0.9cm}
  >{\centering\arraybackslash}p{0.9cm}
  >{\centering\arraybackslash}p{1.1cm}
}
\toprule
\textbf{System} &
\textbf{Storage} &
\rotatebox{70}{\textbf{Block addr.}} &
\rotatebox{70}{\textbf{Dyn.\ order}} &
\rotatebox{70}{\textbf{Agent write}} &
\rotatebox{70}{\textbf{Provenance}} &
\rotatebox{70}{\textbf{Zero I/O}} &
\rotatebox{70}{\textbf{Pos.-aware}} &
\rotatebox{70}{\textbf{Traj.\ learn.}} &
\rotatebox{70}{\textbf{No collapse}} &
\rotatebox{70}{\textbf{CPU compat.}} &
\rotatebox{70}{\textbf{Eviction}} &
\rotatebox{70}{\textbf{No grads}} &
\rotatebox{70}{\textbf{Orch.\ mem.\ mgr.}} \\
\midrule
ReAct~\cite{yao2023react}            & None (static)  & \symnone & \symnone & \symnone & \symnone & \symfull & \symnone & \symnone & \symnone & \symfull & \symnone & \symfull & \symnone \\
MemGPT/Letta~\cite{packer2023memgpt} & DB             & \symfull & \symnone & \symfull & \symnone & \symnone & \symnone & \symnone & \sympart & \sympart & \sympart & \symfull & \sympart \\
AutoGen~\cite{wu2023autogen}          & None (msgs)    & \symnone & \symnone & \symnone & \symnone & \symfull & \symnone & \symnone & \symnone & \symfull & \symnone & \symfull & \symnone \\
Voyager~\cite{wang2023voyager}        & F (code)       & \sympart & \symnone & \symfull & \symnone & \symnone & \symnone & \sympart & \symnone & \symnone & \symnone & \symfull & \symnone \\
ExpeL~\cite{zhao2024expel}            & F + pool       & \symnone & \symnone & \sympart & \symnone & \symnone & \symnone & \symfull & \symnone & \sympart & \symnone & \symfull & \symnone \\
SCULPT~\cite{kumar2025sculpt}         & S (monolithic) & \symnone & \symnone & \symnone & \symnone & \symfull & \symnone & \symnone & \symnone & \symfull & \symnone & \symfull & \symnone \\
SKILL.md~\cite{xu2026skills}        & F (markdown)   & \sympart & \symnone & \symnone & \symnone & \symnone & \symnone & \symnone & \symnone & \symfull & \symnone & \symfull & \symnone \\
A-MEM~\cite{xu2025amem}               & Vector DB      & \symfull & \sympart & \symfull & \symnone & \symnone & \symnone & \symfull & \sympart & \symnone & \sympart & \symfull & \symnone \\
ACE~\cite{zhang2025ace}               & F + IC         & \sympart & \symnone & \symfull & \symnone & \symnone & \symnone & \symnone & \symfull & \symfull & \sympart & \symfull & \symnone \\
\midrule
\textbf{RAMPART}               & \textbf{RAM}   & \symfull & \symfull & \symfull & \symfull & \symfull & \symfull & \symfull & \symfull & \symfull & \symfull & \symfull & \symfull \\
\bottomrule
\end{tabular}
 
\vspace{6pt}
\noindent Column definitions.
\textit{Block addr.}: blocks are individually addressable by a stable identifier.
\textit{Dyn.\ order}: ordering is mutable at runtime.
\textit{Agent write}: the agent can append new instruction blocks at runtime.
\textit{Provenance}: seed, agent, and orchestrator blocks are distinguishably tagged.
\textit{Zero I/O}: no file or database access occurs after session initialization.
\textit{Pos.-aware}: compilation places blocks at attention-favoured positions informed by position-sensitivity research.
\textit{Traj.\ learn.}: experience from task trajectories improves future runs without parameter updates.
\textit{No collapse}: the system prevents iterative erosion of domain-specific knowledge \cite{zhang2025ace}.
\textit{CPU compat.}: the system is practically deployable on consumer CPU hardware with quantized models.
\textit{Eviction}: blocks can be removed based on priority, age, or token budget.
\textit{No grads}: the system operates entirely at inference time without gradient-based parameter updates.
\textit{Orch.\ mem.\ mgr.}: the orchestrator directly writes into and promotes within subagent registries as a zero-token coordination primitive.
\end{sidewaystable}

\section{Conclusion}
\label{sec:conclusion}

RAMPART introduces context assembly as an explicit runtime operation, treating instruction ordering, selective inclusion, and eviction as composable primitives rather than static authoring decisions. The central claim is that what an agent's context contains, and where each element sits within it, should be programmable at the moment of compilation rather than fixed at session load. The probes show, within this controlled setting, that position matters, that grouping the critical block with its content-adjacent neighbours and ordering the group as a unit produces measurable lift at positions where single-block placement fails, and that gating before promotion reduces token cost substantially while preserving most of the position advantage. Each of these results follows from the same underlying mechanism, the compile step treated as a first-class programmable operation rather than a concatenation pass.

The content-priming effect appears at the same absolute positions across five instruction-tuned models from three independent labs, with magnitude graded by model strength on the benchmark and invisible at both ends of the strength range. This suggests an operational protocol of running diagnostic sweeps on a weaker model to surface priming neighbours and then deploying the resulting cluster arrangement on the production model. The same-family scale comparison shows how weak models can perform like larger ones in the regions where the larger model has limited attention to allocate. 

Removing a tool schema block from the registry before compilation prevents the model from invoking the tool not because it was instructed not to, but because the schema never appeared in its context. The model encounters no refusal and no policy instruction and has no knowledge the tool existed. Deletion occurs prior to model exposure, which eliminates the refusal and retention dynamics that arise when a model is instructed to ignore content it has already read. This is a stronger access control property for schema-mediated tool invocation than any instruction-based approach can provide, and it follows directly from the compile-time architecture. A coordinating registry can revoke a subagent's access to a tool between inference calls with no round-trip, no message, and no trust assumption. The revocation takes effect the moment the next compile call runs. The same mechanism generalises to any access boundary that must take effect immediately and completely. Privacy constraints, mandate changes, or scope restrictions imposed by a supervisor registry can be enforced by eviction rather than by instruction.
 
When registry instances hold direct references to one another, agents can exchange constraint blocks, promote proposals, and evict rejected content without any inference call for the coordination itself. A workshop-style multi-agent workflow that would consume hundreds of thousands of tokens under message-passing reduces to zero coordination tokens under shared-registry coordination, because the coordination layer is memory assignment rather than inference. This is not a marginal optimisation but a structural change in how multi-agent workflows can be composed. The topology of who constrains whom is expressed in provenance tagging and non-evictable flags rather than in a messaging protocol.

In structural forgetting, the agent that receives a compiled context without an evicted block has no signal that anything was removed. It produces no refusal and no free-text reference to the absent content. Forgetting in RAMPART is not a content management decision that happens during retrieval. It is a property of what the compilation step produces, and it operates before the model ever examines the context. This matters wherever an agent must lose access to information in a way that cannot be overridden by subsequent instructions.
 
RAMPART is released as a standalone Python library with no inference-backend dependency. The primitives are composable and individually cost zero prompt tokens, making the overhead of adoption the cost of the compile call itself. Existing SKILL.md and CLAUDE.md files import without modification.
\bibliographystyle{unsrtnat}
\bibliography{rampart}

\end{document}